\pdfoutput=1
\documentclass[11pt]{article}

\usepackage[]{ACL2023}

\usepackage{times}
\usepackage{latexsym}
\usepackage{graphicx}
\usepackage{hyperref}
\usepackage{subcaption}
\usepackage{booktabs}
\usepackage[frozencache=true,cachedir=minted-cache]{minted}
\usepackage[T1]{fontenc}
\usepackage[utf8]{inputenc}

\usepackage{microtype}

\usepackage{inconsolata}

\title{NLP at UC Santa Cruz at SemEval-2024 Task 5: Legal Answer Validation using Few-Shot Multi-Choice QA}

\author{Anish Pahilajani$^*$ \and Samyak R Jain$^*$ \and Devasha Trivedi$^*$ \\ Baskin School of Engineering \\ University of California, Santa Cruz \\ \texttt{\{apahilaj,srajeshj,detrived\}@ucsc.edu}}

\begin{document}
\maketitle
\begin{NoHyper}
\def\thefootnote{*}\footnotetext{Equal contribution}
\end{NoHyper}

\begin{abstract}
This paper presents our submission to the SemEval 2024 Task 5: The Legal Argument Reasoning Task in Civil Procedure. We present two approaches to solving the task of legal answer validation, given an introduction to the case, a question and an answer candidate. Firstly, we fine-tuned pre-trained BERT-based models and found that models trained on domain knowledge perform better. Secondly, we performed few-shot prompting on GPT models and found that reformulating the answer validation task to be a multiple-choice QA task remarkably improves the performance of the model. Our best submission is a BERT-based model that achieved the 7th place out of 20. 

\end{abstract}

\section{Introduction}

The field of Natural Language Processing (NLP) has made significant strides in understanding and generating human language.\;Yet, specialized fields such as legal reasoning within the sphere of civil procedure pose distinct challenges. These challenges stem from the intricate nature of legal texts and the requisite domain-specific knowledge. In this paper, we present a solution to the problem posed in Semeval 2024 Task 5, which introduces a new NLP task and dataset focused on the U.S. civil procedure. Our approach in this task aims to evaluate the ability of large language models (LLMs) in interpreting and applying legal principles and laws to specific case questions. To support this paper, we have made our codebase publicly available as a GitHub repository.\footnote{Code available here: \href{https://github.com/devashat/UCSC-NLP-SemEval-2024-Task-5/}{https://github.com/devashat/UCSC-NLP-SemEval-2024-Task-5/}} 
This repository contains all our code and instructions how to run it. At the request of the task organizers, we have not included the dataset splits as they are meant to be private. 

The dataset for this task is curated from ``The Glannon Guide to Civil Procedure'' \cite{glannon2019civil}. It comprises a series of legal cases, each with a general introduction, a specific question related to U.S. civil procedure, and a possible answer candidate. For every answer choice, a comprehensive analysis is provided, rationalizing why it is correct or incorrect. A correct answer is labeled with 1, and an incorrect answer is labeled with 0. The dataset is available in English, and its training, validation, and test splits contain 666, 84, and 98 examples respectively.

\section{Related Work}

The domain of legal question-answering systems has witnessed substantial progress, utilizing cutting-edge computational techniques to address the complexities of legal discourses. A prime example of innovation in this field is the LegalBERT system, introduced by Chalkidis et al. (2020), illustrating the enhanced efficacy of models tailored specifically for legal content through the pretraining of BERT \cite{devlin2019bert} on legal documents \cite{chalkidis-etal-2020-legal}. In addition, \citet{khazaeli2021} achieved a notable breakthrough by developing a flexible legal question-answering system that goes beyond conventional query patterns, incorporating sparse vector search with a BERT-based re-ranking process. The expansion of legal corpora, notably with the Casehold corpus by \citet{zheng2021does}, marked a significant stride forward. This work employed language models for legal analysis, setting a comprehensive standard for measuring model effectiveness in legal reasoning tasks using a dataset based on U.S. court cases. Furthermore, the creation of targeted NLP tasks has played a crucial role in the assessment of models and systems in this field. An important example is the task introduced by \citet{bongard2022legal}, which is the focus of this paper.

These works collectively underscore the diverse methodologies and technological advancements employed in legal question answering. Each of these contributions brings unique insights and solutions, paving the way for more sophisticated and efficient legal question-answering systems in the future.

\section{System Overview}

Our approach at creating a system entailed utilizing few-shot prompting on OpenAI's GPT-3.5 \cite{openai2023gpt35} and GPT-4 \cite{openai2023GPT4}. We devised a script with a prompt, two examples of the expected model input and desired output, and used an altered version of the task dataset as our input queries for this system. We experimented with a few prompts for both GPT models, trying to figure out what gave us the best results. Following the guidance outlined in \citet{bsharat2024principled} and \citet{white2023prompt}, we decided to structure our prompt in the following manner: 

\begin{itemize}
    \item A system instruction that describes to the model the structure of our data, the input it will receive, and what the model should return,
    \item Two examples from the training dataset, one with a correct answer prediction and one with an incorrect answer prediction,
    \item The dataset containing our questions and answer candidates.
\end{itemize}

Additionally, we also altered the dataset from a binary classification format to a multi-choice QA format. Rather than presenting individual question-answer pairs, each question was now accompanied by the entire set of potential answer choices. \autoref{fig:dataset_format} demonstrates a visual example of this restructuring. As can be inferred from \autoref{fig:dataset_format}, the binary classification format of the dataset requires the system to label each answer choice as 0 or 1, whereas the multi-choice format requires the system to return one single answer prediction for each question. We chose to convert the dataset in this way to address an issue we found in the experimental phase of our work. This issue is elaborated upon in \autoref{sec:experiments}.

A small note on our multi-choice QA format, we added an additional option ``None of the Above'' because in some cases, the training or the validation data had all incorrect answer choices listed for a given question. This was done to make sure that the model would not be forced to pick between multiple incorrect answers.

\begin{figure}[h!] {\small
\textbf{Binary Classification Format}
\begin{minted}{text}
Question: 
<Question>

Context:
<Explanation>

Choice:
<Answer Candidate 1>

Question: 
<Question>

Context:
<Explanation>

Choice:
<Answer Candidate 2>
\end{minted}

\textbf{Multi-Choice QA Format}
\begin{minted}{text}
Question:
<Question>

Context:
<Explanation>

Choices:
{0: <Answer Candidate 1>, 
1: <Answer Candidate 2>, 
2: <None of the Above>}
\end{minted}
}
\caption{Difference between the original dataset format and our restructuring}
\label{fig:dataset_format}
\end{figure}

\autoref{fig:sys_instructions} shows our best performing system instructions for the binary classification format of the dataset and the multi-choice QA format of the dataset. We ran experiments on both dataset formats to compare system performance, and \autoref{sec:results} contains our results for evaluation metrics.

\begin{figure}[h!]
{\small
\textbf{Multi-Choice System Instruction:}
\begin{minted}{text}
You are an AI legal expert with 
expertise in U.S. Civil Procedure 
and U.S. Civil Law, known for your 
strong reasoning abilities. Your 
task is to answer a Multiple 
Choice Question in the legal domain. 
Choose an answer only if you are 
very confident, otherwise, select 
"None of The Above."

You will be provided with:
1. question: A legal question
2. context: Additional context for 
better understanding
3. choices: Multiple answer candidates

Your response should be a JSON with two 
keys: "correct_answer" and "reasoning." 
Place the correct answer exactly as 
provided in the "correct_answer" key. 
Provide a detailed explanation of your 
reasoning in the "reasoning" key. Do 
not add or remove any other text.

Your goal is to ensure accurate 
answers and thorough reasoning.
\end{minted}

\textbf{Binary Classification System Instruction:}
\begin{minted}{text}
You are an AI legal expert with 
expertise in U.S. Civil Procedure 
and U.S. Civil Law, known for your 
strong reasoning abilities. Your 
task is to answer a question in 
the legal domain.

You will be provided with:

1. question: A legal question
2. context: Additional context for 
better understanding
3. answer candidate: an answer candidate 
that can be either correct or incorrect

Your response should be a string with 
length 1. You will be classifying a 
correct answer as 1, and an 
incorrect answer as 0.

Your goal is to ensure accurate 
answers and thorough reasoning.
\end{minted}
}
\caption{System Instructions for Both Dataset Formats}
\label{fig:sys_instructions}
\end{figure}

\section{Experiments}\label{sec:experiments}

\subsection{Finetuning with BERT}\label{sec:bert_exp}

To establish a solid baseline, we opted to fine-tune various BERT models. This process involved inputting both the question and its corresponding answer into the model, with the goal of generating an output label of either 0 or 1. We trained our model on the data for 500 epochs before having it predict. In the predictions we observed a propensity for both BERT models to disproportionately favor the 0 label, a phenomenon likely stemming from the dataset's natural imbalance due to it being formatted for binary classification. Since the source material for the dataset is in a multiple choice format, there are inherently more answers with the 0 label than answers with the 1 label, and a predictive model would tend to prefer the majority label \cite{tanha2020boosting}. After noticing this issue, we tried experimenting with altering the dataset.

\subsection{Data Augmentation}\label{sec:dataset_exp}

To address the challenge of our model's tendency to overfit on the 0 label, we explored incorporating the Casehold corpus into our dataset. Casehold \cite{zheng2021does}, a rich legal corpus derived from the Harvard case law collection and spanning from 1965 to the present, was initially formatted for multi-label use, offering a wealth of potential answers for each question. Despite our efforts to adapt this corpus into a binary format to align with the organizers' dataset format, we encountered persistent overfitting issues, leading us to believe that trying the balance the dataset would not yield any productive results. 

Subsequently, we reverted to using solely the task dataset and refined our approach by integrating each question's text with its corresponding explanation to provide more context. This was done with the hope that our model would use the additional input to steer its prediction in the correct direction. However, this addition faced a technical bottleneck due to the 512-token limit inherent in the BERT models, prompting us to investigate alternative large language models (LLMs) that could handle the larger input size. We decided to explore two options. The first was finetuning Longformer because it uses windowed attention and can handle longer context lengths \cite{beltagy2020longformer}. The second was exploring few-shot prompting with GPT-3.5 and GPT-4 as we could run further experiments also comparing how the GPT models do with both formats of the dataset, if the overfitting issue would persist or if one dataset format would outperform the other. We also wanted to try few-shot prompting as \citet{brown2020language} found it to be a better approach for QA tasks than finetuning.

\subsection{Finetuning Longformer}

Finetuning Longformer helped us resolve the context length limit that we ran into with BERT, but it did not yield better results. Considering that in our experiments with BERT we found LegalBERT to be the better performing model (see \autoref{sec:results}), we decided to use a Longformer model with legal context embedded into it. This legal Longformer model is devised by \citet{chalkidis-etal-2023-lexfiles} and is a derivative model of a base RoBERTa trained on the LexFiles corpus \cite{chalkidis-etal-2023-lexfiles}.

Using this legal Longformer model, we were able to incorporate the explanation feature into our input. Our input was explanation, question, answer. We first ran the finetuning code for 100 epochs where we ran into the same problem of overfitting. In fact, our F1 score would not go above 44.37 on the validation set - the model would only predict 0 labels and performed worse than finetuning the BERT models. We then increased the number of epochs to 500, but that also did not show an improvement in F1 scores on the validation set. We believe that Longformer performed worse than LegalBERT due to differences in their pretraining corpora.

\subsection{Few-Shot Prompting with GPT-3.5 and GPT-4}\label{sec:gpt_exp}

Our experimentation with GPT-3.5 and GPT-4, through few-shot prompting, offered promising directions. Notably, this method enabled us to effectively incorporate even the analysis feature of the dataset within the context limit, achieving an impressive F1 score of 90 on the validation set. Despite this success, the approach did not consistently extend to the test set, suggesting that using analysis to predict correct answers has its limitations, as the test set inputs lacked the feature.

In our final strategy to mitigate the dataset's imbalance, we shifted from binary to multi-choice classification, allowing for a more nuanced model assessment. This change meant our model now aimed to identify the correct answer from a set of options, rather than simply labeling each answer as 0 or 1. Reapplying few-shot prompting to GPT-3.5 and GPT-4, with the dataset's adjusted format, led to our most improved performance on the dataset.

For the prompting experiments, we used the OpenAI API. We ran our prompting code for 3 epochs, and did not alter any other hyperparameters. 

\subsection{Rule-based Algorithm Application} 

After successfully implementing a prediction system, we increased our F1 score and accuracy by applying a rule-based algorithm tailored to the characteristics of each dataset. Recognizing the inherent imbalance within the datasets, we devised a strategy where if all answers to a question were labeled as 0 in the training and validation sets, then the answer for said question in the test set was presumed to be labeled as 1. Conversely, if there were any correct answers in the training or validation sets, the data entries with the corresponding question were considered incorrect in the test set. This adjustment allowed us to enhance our performance metrics significantly for the baseline BERT models and the GPT-3.5 and GPT-4 predictions, giving us the metrics outlined in \autoref{tab:RBBERT} and \autoref{tab:RBGPT} for the test dataset. We only utilized this technique for the competition part of the task, as we wanted to see how high we could score. We did not submit our predictions with the GPT models.

\section{Results}\label{sec:results}

Our submission to the SemEval task ranked 7th out of 20 on the competition leaderboard. What we submitted to the task competition was our best performing finetuned BERT model after applying the rule-based algorithm. This was not our best method, as we were able to achieve higher metrics through subsequent experimentation. Our best results overall can be seen in \autoref{tab:RBGPT}, which stem from few-shot prompting on GPT models using the multi-choice QA format of the dataset, and then applying the rule-based algorithm to the predictions generated.

\autoref{tab:BERT} presents our F1 score and accuracy across the two BERT models we chose to finetune. Our experiments not only aligned with but also surpassed the benchmarks established by the task organizers \cite{bongard2022legal}, achieving a 0.2 increase in F1 score by merely utilizing the question and answer features in the input coupled with our fine-tuning approach. When comparing the baseline results from \citet{bongard2022legal} with our own, we found that our question, answer input alone had a similar score to their input that also utilized the explanation feature. They achieved a 65.73 F1 score, whereas our submission to the task competition achieved a 65.99 F1 score as shown in \autoref{tab:RBBERT}.

Our best results without utilizing the rule based algorithm came from few-shot prompting with GPT models using the multi-choice QA format of the dataset. \autoref{tab:MCGPT} shows these result metrics, while \autoref{tab:BCGPT} shows the metrics of few-shot prompting using the binary classification format of the dataset for comparison.

\begin{table}[ht]
    \centering
    \begin{tabular}{p{3.5cm}cc}
    \toprule
    Model & F1 Score & Accuracy \\
    \midrule
    BERT & 57.56 & \textbf{73.47} \\
    LegalBERT & \textbf{63.27}  & 72.45 \\
    \bottomrule
    \end{tabular}
    \caption{Baseline fine-tuned BERT models and their performance on test set.}
    \label{tab:BERT}
\end{table}

\begin{table}[ht]
    \centering
    \begin{tabular}{lcc}
    \toprule
    Model & Test F1 Score & Test Accuracy \\
    \midrule
    GPT-4 & \textbf{71.70} & \textbf{80.61} \\
    GPT-3.5 & 62.21 & 72.45 \\
    \bottomrule
    \end{tabular}
    \caption{Results of Multi-Choice QA Few Shot Prompting on GPT Models.}
    \label{tab:MCGPT}
\end{table}

\begin{table}[ht]
    \centering
    \begin{tabular}{lcc}
    \toprule
    Model & Test F1 Score & Test Accuracy \\
    \midrule
    GPT-4 & \textbf{68.77} & \textbf{73.47} \\
    GPT-3.5 & 48.64 & 48.98 \\
    \bottomrule
    \end{tabular}
    \caption{Results of Binary Classification Few Shot Prompting on GPT Models.}
    \label{tab:BCGPT}
\end{table}

\begin{table}[ht]
    \centering
    \begin{tabular}{lcc}
    \toprule
    Model & F1 Score & Accuracy \\
    \midrule
    BERT & 59.99 & 74.49 \\
    LegalBERT & \textbf{65.99}  & \textbf{74.49} \\
    \bottomrule
    \end{tabular}
    \caption{Results of combining a rule-based algorithm with finetuning on BERT models.}
    \label{tab:RBBERT}
\end{table}

\begin{table}[ht]
    \centering
    \begin{tabular}{lcc}
    \toprule
    Model & Test F1 Score & Test Accuracy \\
    \midrule
    GPT-4 & \textbf{74.68} & \textbf{82.65} \\
    GPT-3.5 & 64.13 & 73.47 \\
    \bottomrule
    \end{tabular}
    \caption{Results of combining a rule-based algorithm with multi-choice few shot prompting on GPT models}
    \label{tab:RBGPT}
\end{table}

\section{Conclusion}

Our investigation into the application of Large Language Models (LLMs) in the domain of legal reasoning for civil procedure, as a contribution to SemEval 2024 Task 5, has led us to several significant insights. These insights not only highlight the capabilities and limitations of current AI technologies in legal applications but also chart a course for future advancements in this intriguing intersection of technology and jurisprudence.

\subsection{Best system} 

Our research identified that the application of multi-choice QA few-shot prompting on GPT-4 was the most effective method, achieving an F1 score of 71.70 and an accuracy of 80.61 on the test dataset. A significant insight from our experiments is the inherent limitation encountered with BERT models, notably their 512-token context length constraint. This limitation poses a unique challenge in legal reasoning tasks, where the richness and complexity of legal texts often necessitate a comprehensive contextual understanding that exceeds the input capacity of traditional models. By successfully navigating these constraints with GPT-4's advanced capabilities, our approach demonstrates the benefits of leveraging the more flexible and expansive context handling offered by newer generation models to effectively process and interpret dense legal information.

\subsection{Impact of analysis feature} 

The inclusion of an analysis feature significantly improved LLM performance during the fine-tuning process on both the training and validation datasets. However, the anticipated benefits of this feature did not extend to the test dataset, likely due to differences in input structure between the training/validation and test phases. This suggests a potential overfitting problem, indicating that while models may become adept at recognizing patterns in training data, they may not necessarily understand the fundamental legal reasoning principles underlying the data.

\subsection{Format of Dataset} 

The imbalance of the dataset, coupled with the fact that it was primarily sourced from a single textbook, introduced a challenge in preventing models from exploiting its predictable structure. To foster more rigorous and analytically profound datasets in this research domain, we propose diversifying the sources of dataset content. Additionally, we suggest that future datasets should challenge models to not only select the correct answer but also to generate the reasoning behind their choices. This method could provide a use case for the analysis feature, promoting a deeper understanding and application of legal principles, leveraging the full potential of Generative AI in legal reasoning.

\subsection{Future Work}

Exploring the integration of specific laws or precedents as a form of analysis presents an intriguing direction for enhancing the capabilities of Large Language Models (LLMs) in legal reasoning tasks. This approach deviates from the current format of analysis; explaining why an answer choice is correct or incorrect. Instead, it involves presenting the LLM with the relevant legal principles or statutes directly related to the question input. The model is then tasked with interpreting these legal documents to deduce the correct answer based on the law's stipulations.

Such a methodology could foster the model reaching a deeper level of engagement with the material, as it not only challenges the model to grasp the nuances of legal language but also might help improve the model's ability to generalize from the principles of law to the specifics of individual cases, potentially leading to more accurate and legally sound predictions. As such, future work could involve curating or enhancing existing datasets to include these legal references, alongside developing model architectures and training methodologies that are adept at handling such complex, text-based inputs.

\paragraph{Acknowledgements} We would like to express our gratitude to Professor Ian Lane, Professor Jeffrey Flanigan, Nilay Patel, and Jeshwanth Bheemanpally from University of California, Santa Cruz. Their comments, insights, and feedback helped us along the process of participating in this task and writing this paper.

\bibliographystyle{apalike}
\bibliography{references}

\end{document}